\documentclass{article} 
\usepackage{nips15submit_e,times}
\usepackage{hyperref}
\usepackage{graphicx}
\usepackage{url}
\usepackage{caption}
\usepackage{subfigure}

\usepackage{amsmath}

\usepackage{listings}
\usepackage{xcolor}

\definecolor{codegreen}{rgb}{0,0.6,0}
\definecolor{codegray}{rgb}{0.5,0.5,0.5}
\definecolor{codepurple}{rgb}{0.58,0,0.82}
\definecolor{backcolour}{rgb}{0.95,0.95,0.92}

\lstdefinestyle{mystyle}{
    backgroundcolor=\color{backcolour},   
    commentstyle=\color{codegreen},
    keywordstyle=\color{magenta},
    numberstyle=\tiny\color{codegray},
    stringstyle=\color{codepurple},
    basicstyle=\ttfamily\footnotesize,
    breakatwhitespace=false,         
    breaklines=true,                 
    captionpos=b,                    
    keepspaces=true,                 
    numbers=left,                    
    numbersep=5pt,                  
    showspaces=false,                
    showstringspaces=false,
    showtabs=false,                  
    tabsize=2
}

\title{Pixel to policy: DQN Encoders for \\ within \& cross-game reinforcement learning}

\author{
Sourabh Prakash\\
Halıcıoğlu Data Science Institute\\
University of California, San Diego\\
La Jolla, CA 92092 USA \\
\texttt{soprakash@ucsd.edu} \\
\And
Priyanshi Shah\\
Halıcıoğlu Data Science Institute\\
University of California, San Diego\\
La Jolla, CA 92092 USA \\
\texttt{prs003@ucsd.edu} \\
\And
Ashrya Agrawal\\
Computer Science and Engineering\\
University of California, San Diego\\
La Jolla, CA 92092 USA \\
\texttt{asagrawal@ucsd.edu} \\
}

\nipsfinalcopy 

\begin{document}

\maketitle





\begin{abstract}
    \item Reinforcement Learning can be applied to various tasks, and  environments. Many of these environments have a similar shared structure, which can be exploited to improve RL performance on other tasks. Transfer learning can be used to take advantage of this shared structure, by learning policies that are transferable across different tasks and environments and can lead to more efficient learning as well as improved performance on a wide range of tasks. This work explores as well as compares the performance between RL models being trained from the scratch and on different approaches of transfer learning.  Additionally, the study explores the performance of a model trained on multiple game environments, with the goal of developing a universal game-playing agent as well as transfer learning a pre-trained encoder using DQN, and training it on the same game or a different game. Our DQN model achieves a mean episode reward of 46.16 which even beats the human-level performance with merely 20k episodes which is significantly lower than deepmind's 1M episodes. The achieved mean rewards of 533.42 and 402.17 on the Assault and Space Invader environments respectively, represent noteworthy performance on these challenging environments.
    
\end{abstract}
\section{Introduction}
Reinforcement learning is a popular approach for training agents to make decisions by maximizing their reward through a learned policy. However, training a reinforcement learning model can take a significant amount of time (7-8 days on GPU) due to the need for large datasets and multiple episodes, resulting in months of training and fine-tuning. The objective of our paper is to reduce the training time of reinforcement learning models to hours from days, specifically in the context of understanding the shared structure of Atari games. 

To test our hypothesis, we first employed Deep Q-Network (DQN), a popular approach for value-based reinforcement learning, but found that it has a significant convergence time, taking up to seven days to train a satisfactory model. To overcome this challenge, we implemented transfer learning by utilizing a pre-trained encoder of a similar Atari game \cite{gogianu2022agents} and applying it to a new environment with nearly-similar gameplay. The aim is  to leverage the knowledge gained from training on one game to improve performance on another. We devised three distinct methods for this process,which we elaborate on in detail in the upcoming sections. The paper also explores towards developing a universal game-playing agent, that can perform reasonably well on games with similar action space. 

Through this exploration, we aim to improve the efficiency of our reinforcement learning algorithms and obtain better results in understanding the shared environment structure of Atari games.

Overall, the objective of this paper is to reduce the time taken to train reinforcement learning models by utilizing transfer learning and developing universal agent , while also addressing other challenges like credit assignment problem, exploration-exploitation trade-off and lack of generalization.  Our experiments demonstrate that these methods can significantly reduce the training time and improve the performance of reinforcement learning models in Atari games. This can be expanded to have potential applications in developing more efficient and effective reinforcement learning algorithms for a wide range of real-world applications. 


\section{Related Work}
In this rapidly advancing era, Reinforcement learning(RL) {\cite{sutton2018reinforcement}} has seen tremendous increase with various algorithms and techniques being introduced along with a wide variety of applications. One of the most famous works, Reinforcement Learning: An Introduction\cite{sutton2018reinforcement}, provides a detailed  overview to the field of reinforcement learning, starting from its historical context, foundational papers, concepts, algorithms and domain related applications. 

Deep Q-Networks(DQN), one of the most famous algorithms used in RL combines Q learning and deep neural networks. The paper \cite{Mnih2015HumanlevelCT} achieved more than human level performance on many Atari games. This was an advancement of one the previous papers, Playing Atari with Deep RL \cite{mnih2013playing}for high dimensional input spaces. 

Another work on Transfer learning for reinforcement learning\cite{taylor2009transfer} domains gives an extensive survey of transfer learning techniques for including model-based and model-free RL approaches and highlights the challenges and opportunities in this area. 

Furthermore, the Proximal Policy Optimization (PPO) was introduced as a new family of RL algorithms \cite{schulman2017proximal} that iteratively updates the policy based on samples collected from the environment. combines the simplicity and efficiency of policy gradient methods with the stability and reliability of value-based methods. PPO has achieved state-of-the-art performance on a wide range of benchmark tasks, including Atari games and continuous control tasks. The algorithm has also been extended to handle more complex scenarios, such as multi-agent reinforcement learning.

\section{Datasets}
The OpenAI Gym Atari games dataset is a collection of gameplay data from a set of Atari games, generated using the OpenAI Gym framework. The dataset is designed to be used as a benchmark for evaluating the performance of reinforcement learning algorithms on Atari games, and is widely used in research on deep reinforcement learning.

The dataset includes gameplay data from a set of 60 Atari games, including classic games such as Pong, Space Invaders, and Pac-Man. For each game, the dataset includes a set of preprocessed game frames and associated action and reward data, which can be used to train and evaluate reinforcement learning algorithms.

The game frames in the dataset are preprocessed to reduce the complexity of the input data and make it easier for reinforcement learning algorithms to learn from the data. These are much more well-maintained and also present interesting possibilities. These are a collection of 2600 environments many of which have the same state space - RGB image of same dimensions(84 x 84 x 3). Multiple groups of similar tasks can be found in Gym Atari games dataset.

\subsection{Preprocessing of input frames}
Preprocessing is an important step in reinforcement learning (RL), as it can greatly affect the learning performance and stability of the algorithm. Our pipeline uses a series of preprocessing steps to the raw images, including resizing the images to a smaller size(84 x 84 x 3), converting the images to grayscale, and normalizing the pixel values to the range [0, 1]. The resizing step is performed to reduce the dimensionality of the observation space, making it easier for the RL agent to learn a policy. The grayscale conversion step reduces the computation cost of training by reducing the number of channels required in the input observation, while preserving important visual features. The normalization step ensures that the pixel values are in the same range across all images, making it easier for the RL agent to learn a consistent policy. Additionally, we apply frame skipping, which means that every k-th frame is used as an observation and the actions are repeated for k-1 frames. This can help reduce computation time and stabilize the learning process by reducing the correlation between consecutive observations. The preprocessing steps applied by the AtariPreprocessing wrapper can reduce the dimensionality of the observation space and provide a more informative and stable representation of the game state to the RL agent. Overall, the AtariPreprocessing wrapper is a powerful tool for preprocessing raw pixel observations in Atari games, and can greatly improve the efficiency and stability of RL algorithms.

\subsection{Frame Stacking}
In this paper, we use frame stacking, a widely used technique in reinforcement learning (RL), to improve the efficiency and effectiveness of our deep RL experiments involving Atari games. We employ the Frame stacking wrapper in the Gym library. By stacking multiple consecutive frames together to create a single observation, we are able to provide a more informative representation of the game state to the RL agent. This technique helps the RL agent better capture the dynamics of the game, making it easier to learn a stable policy. Overall, frame stacking is a powerful technique for improving the efficiency and effectiveness of RL algorithms, and we noted its effectiveness in our experiments involving Atari games. In our breakout game training, when frame stacking wasn't employed, the model was not performing well at sensing the direction of the ball, but with frame stacking, it got better at capturing the mechanics of the game like direction, velocity and acceleration.

\section{Methods}
The objective of our paper is to utilize reinforcement learning techniques for understanding the shared structure of the environment in Atari games. We begin our experiments by employing Deep Q-Network (DQN)\cite{pytorch_dqn_tutorial}, which is a popular approach for value-based reinforcement learning. However, we noticed that this method has a significant convergence time, and it took us up to seven days to train a satisfactory model.

To overcome this challenge, we implemented transfer learning by utilizing a pre-trained encoder of a similar Atari game and applying it to a new environment with similar gameplay. We devised two distinct methods for this process, which we elaborate on in detail in the upcoming sections.

Our DQN architecture consists of two main components: a feature encoder and fully connected layers. The feature extractor is a sequential module that includes three convolutional layers with ReLU activation functions. The first layer has 32 filters and a kernel size of 8 with a stride of 4, the second layer has 64 filters and a kernel size of 4 with a stride of 2, and the third layer has 64 filters and a kernel size of 3 with a stride of 1. The policy network consists of fully connected layers - two linear layers with ReLU activation functions. The first linear layer has 64 x 7 x 7 input features (corresponding to the output of the last convolutional layer) and outputs 512 features. The second linear layer then takes the 512 features and produces the final output size, which is characterized by the action space of the model.

\subsection{DQN Model- from scratch}

Deep Q-Networks (DQNs) are a popular type of deep reinforcement learning (RL) algorithm that has been successfully applied to a variety of RL problems, including Atari games. DQNs use a deep neural network to approximate the optimal Q-function, which is the expected total reward for taking a given action from a given state and following the optimal policy thereafter. The optimal Q-function satisfies the Bellman equation, which is given by:

\begin{equation}
Q(s,a) = E[r + \gamma \max_{a'} Q(s',a') | s,a]
\end{equation}

where $s$ is the current state, $a$ is the action taken in that state, $r$ is the reward received, $\gamma$ is the discount factor, $s'$ is the next state, and $a'$ is the next action.

To learn the Q-function, our DQN uses a replay buffer, which stores experiences $(s,a,r,s',\text{done})$ in a FIFO queue. During training, a minibatch of experiences is sampled from the replay buffer and used to update the Q-network parameters. This approach helps to reduce the correlation between consecutive experiences and improves the stability of the learning process.

In addition, we use a policy-target network architecture, which consists of two neural networks: a Q-network and a target network. The Q-network is used to select actions during training, while the target network is used to estimate the optimal Q-function. The target network is updated periodically with the parameters from the Q-network using a soft-update rule:


\begin{equation}
\theta_{\text{target}} = \tau \theta_{\text{Q}} + (1-\tau) \theta_{\text{target}}
\end{equation}

where $\theta_{\text{Q}}$ and $\theta_{\text{target}}$ are the parameters of the Q-network and target network, respectively, and $\tau$ is a hyperparameter that controls the rate of update. This soft-update rule helps to stabilize the learning process and prevent overfitting.

Overall, DQNs are a powerful tool for deep RL and have been successfully applied to a wide range of problems. By using a replay buffer, policy-target network architecture, and soft-update rule, DQNs can learn to approximate the optimal Q-function and achieve state-of-the-art performance on Atari games and other RL tasks


\subsection{Asynchronous Parallel Execution for improving sample generation}
Reinforcement learning algorithms often require large amounts of data to learn optimal policies, which can be a time-consuming process. Asynchronous parallel execution is one approach that has been shown to improve the efficiency of sample generation for reinforcement learning algorithms. In this work, we employ the gym library's vector module to create a vectorized environment that allows for asynchronous parallel execution of multiple instances of an environment. This enables us to generate multiple trajectories in parallel, allowing the agent to explore a larger portion of the state space and learn more efficiently.

To ensure that the asynchronous execution does not affect the stability of the learning algorithm, we use a replay buffer to store the samples generated by the asynchronous execution. We use a variant of the prioritized experience replay algorithm to ensure that the replay buffer contains diverse and informative samples. Overall, the use of asynchronous parallel execution with gym is a powerful tool for improving the efficiency of sample generation in reinforcement learning. Our results demonstrate that this approach can significantly speed up the learning process, allowing us to explore larger portions of the state space and learn optimal policies more efficiently

\subsection{Reinforcement Learning - Transfer learning}
Transfer learning is a machine learning technique to train a model on one task so as to leverage and improve the performance on a related task. Transfer learning in RL refers to the practice of using a pre-trained agent as a starting point for learning a new task. This fine-tuning process involves adjusting the weights of the pre-trained agent's neural network to better adapt to the new task. This helps in situations when there is less availability of data or computational resources and can also help accelerate the learning process by providing a good starting point for learning a new task, while improving the performance of the agent.

In DQN, the most computationally expensive part is typically training the Q-network, which is a type of deep neural network (CNN is this case), due to the need to be trained on a large dataset and for a large number of episodes. Using a pretrained CNN encoder helps in reducing the training time by 20 folds (from 7 days to 7-8 hours). We have employed three distinct methods for implementing transfer learning in the CNN network, which vary according to the initialization of the network's weights and the degree of weight freezing. The CNN network comprises two main components: the feature extractor, which is responsible for extracting features from the input images, and the head network, which takes the extracted features as input and produces the corresponding action to be taken.

The feature extractor consists of three convolutional layers with 32, 64, and 64 filters, and ReLU as the non-linear layer. The head network consists of two linear layers with a ReLU activation function between them. It takes in 3136 input features, outputs 512 features in the first layer, and 6 features in the second layer. It produces Q-values for each of the 6 possible actions in a given state. The overall process remains same for all the three approaches, which takes in (84 x 84) images, converts it into gray scale and creates a stack of 4 images. These images are then subtracted so that the network could extract attributes such as velocity, direction etc. The policy network is updated using the gradient based on the difference between the predicted Q-value and target Q-value, estimated using the target network.

\subsubsection {Within Game Initialization} 

For the first approach to transfer learning using the pretrained CNN encoder, a pre-trained encoder from a similar environment was selected. The weights of the feature extractor were frozen, and a new head-layer consisting of two fully connected neural networks (FCNs) was added to the existing architecture. The model was then fine-tuned on the new task using the frozen feature weights, allowing the FCN weights to be optimized for the new task. SpaceInvaders and Assault environments were used for this experiment. First, the pre-trained encoder from the same environment was used to fine-tune the model and for the second experiment, pretrained weights from Space Invaders environment encoder were used to fine-tune assault and vice-versa. One key difference to note between the two is that the last layer of space invaders had 7 output nodes whereas for SpaceInvaders it was 6 and it resulted in an interesting observation. In short, the results were not symmetrical. The training was done for 400 episodes and resulted in a decent performance.  The hyperparameters used are: BATCH\_SIZE = 128(number of transitions sampled from the replay buffer), GAMMA = 0.99(discount factor), EPS\_START = 0.9(starting value of epsilon) , EPS\_END(final value of epsilon) = 0.05, EPS\_DECAY(rate of exponential decay of epsilon) = 1000, TAU(target network update rate) = 0.005, LR(learning rate) = 1e-4.

\subsubsection {Cross Game Initialization}

For the second approach to transfer learning using the pretrained CNN encoder, a pre-trained encoder from a similar environment was selected. Depending upon how the policy network is initialized, we have two kinds of networks and hence the experiments. For the first one the weights of the feature extractor were frozen, and the policy network was initialized with the weights from the pre-trained policy network(head layer). Basically, the softmax layer of the policy network was changed, and the weights were initialized with pre-trained weights and the policy network was fine-tuned as per the new environment. For the second experiment, we follow the similar procedure and freeze the pre-trained encoder weights. The new policy network (head-layer) was added, without initializing with pre-trained weights and then trained from scratch. 

The hyperparameters used for both the experiments are: BATCH\_SIZE = 128, GAMMA = 0.99, EPS\_START = 0.9, EPS\_END = 0.05, EPS\_DECAY = 1000, TAU = 0.005, LR = 1e-4. This approach helps in settings where one wants to train a reinforcement learning agent but cannot find a pre-trained model for that particular use case. Using the pre-trained weights from a relatively complex but similar environment can lead to much faster convergence in performance.  

\subsubsection {End-To-End Training}
For the third approach to transfer learning in the CNN encoder, a pre-trained encoder from a similar environment was selected. Unlike the previous approaches, both the feature layer and head-layer weights were initialized with weights from the pre-trained encoder, and were not frozen during training. The model was then fine-tuned on the new task using the initialized weights, allowing the weights to be optimized for the new task while retaining the learned representations from the pre-trained encoder. The hyperparameters used are: BATCH\_SIZE = 128, GAMMA = 0.99, EPS\_START = 0.9, EPS\_END = 0.05, EPS\_DECAY = 1000, TAU = 0.005, LR = 1e-4. 

This method explores the feature space of the new environment and tunes the encoder as per the feature space of the environment starting with the weights of the pretrained encoder. This allows the encoder to converge quickly since the initial layers are generic and only the final layers need to be tuned. This would require more episodes to train when compared to frozen encoder but would result in better performance. On the whole, it would still be much faster than the DQN network trained from scratch.  

\subsubsection {Universal Game-Playing Agent}
A universal game-playing agent is a type of artificial intelligence that can learn to play multiple games without any prior knowledge of the game rules. We explore the development of a universal game-playing agent using Deep Q-Network (DQN) reinforcement learning.  To achieve this, we first train the model on a number of similar Atari environments that have the same action space namely, SpaceInvaders, JourneyEscape, Phoenix and test it on DemonAttack. The hyperparameters used are: BATCH\_SIZE = 128, GAMMA = 0.99, EPS\_START = 0.9, EPS\_END = 0.05, EPS\_DECAY = 1000, TAU = 0.005, LR = 1e-4. 

For the reinforcement learning agent, the  is a new unseen environment and the agent was never trained on it, the agent to escape from the enemy's attacks well as attack on the enemy so as to increase rewards by being trained on previous similar environment. During the training the agent's action space was similar and it had to attack to get rewards, and escape the enemy's attack in order to survive. Training on multiple similar environments, generalizes the policy network and since a similar action-reward mapping is needed for the game, we get a decently good performance. 

Though this model does not achieve the state of the art or the human level performance but the experiment results do provide us hope that reinforcement learning agents can be trained towards generalization. We tried a sequential training approach in which we trained the model one each environment one at a time and moved on to the next one. If carefully curated and selected environments are used then the final model can learn to give human level performance on unseen environment as well.

\section{Results}
In this section, we show our experimental results and graphs for all of our methodologies used in this paper. We also show the overall results as seen in Table \ref{tab:overall_results} which shows how with less number of episodes we are able to get decent results. For the Breakout game, we were able to get a better performance than human level. Similarly for Assault and Atari, using our methodology of Transfer Learning as discussed in below subsections, we achieve decent performance in just few episodes. Thus we show how our method can be helpful in various scenarios with less training.
\begin{table}[h]
\centering
\caption{Overall Best Results as compared to Human level}
\label{tab:overall_results}
\begin{tabular}{|c|c|c|c|
}
\hline
\textbf{Game} & \textbf{Model} & \textbf{Rewards} & \textbf{Episodes} \\ \hline
Breakout & Human level & 30.5 & -  \\
 & Our model & 46.16 & 20k \\ \hline
Assault & Human level & 742 & -  \\
 & Our model & 533.42 & 400 \\ \hline
Space Invaders & Human level & 1668.7 & -  \\
 & Our model & 402.17 & 500 \\ \hline
\end{tabular}
\end{table}

\subsection{DQN}
\begin{figure}[ht!]
  \centering
  \subfigure[Rewards vs Episode]{\includegraphics[width=0.45\textwidth]{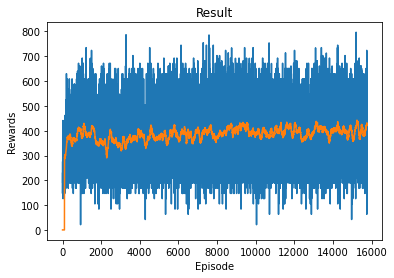}}
  \hfill
  \subfigure[Duration vs Episode]{\includegraphics[width=0.45\textwidth]{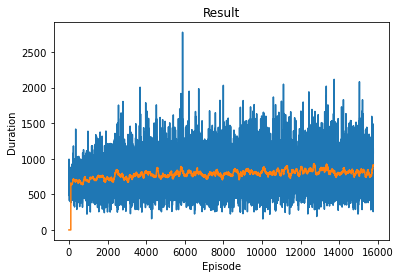}}
  \caption{Assault(Atari): DQN performance when trained from scratch }
  \label{fig:dqn_assault_scratch}
\end{figure}

\begin{figure}[ht!]
  \centering
  \subfigure[Rewards vs Episode]{\includegraphics[width=0.45\textwidth]{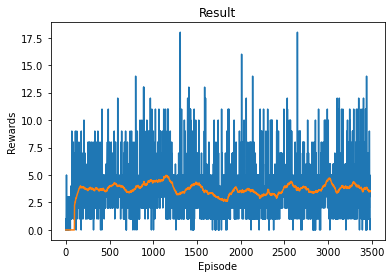}}
  \caption{Breakout(Atari): DQN performance when trained from scratch }
  \label{fig:breakout_scratch_rewards}
\end{figure}

When DQN is trained from scratch, the model shows little to no improvement. This is demonstrated in Figures \ref{fig:dqn_assault_scratch} and \ref{fig:breakout_scratch_rewards}, where the mean reward stays constant at poor reward values. Note that in this figure and the future figures, the blue lines denote episode rewards/loss/duration, whereas the orange line denotes the moving average over 100 values. And uptill 100 values, the moving average value, as depicted in the orange line is set to 0. Table \ref{tab:episode_rewards_assault_scratch} shows that there is no consistent improvement and the mean reward even decreases during training

\begin{table}[ht!]
\centering
\caption{Mean rewards obtained at specific episodes with DQN(from scratch) on Assault(Atari)}
\label{tab:episode_rewards_assault_scratch}
\begin{tabular}{|c|c|c|c|c|c|c|}
\hline
\textbf{Environment} & \textbf{Episode 100} & \textbf{Episode 500} & \textbf{Episode 1000} & \textbf{Episode 5000} & \textbf{Episode 10000} & \textbf{Episode 15000} \\ \hline
Assault & 291.06 & 339.57 & 374.01 & 374.74 & 383.25 & 380.73  \\ \hline
\end{tabular}
\end{table}

\begin{table}[ht!]
\centering
\caption{Mean Rewards obtained at specific episodes for Breakout with DQN(from scratch) on Breakout}
\label{tab:episode_rewards_breakout_scratch}
\begin{tabular}{|c|c|c|c|c|c|}
\hline
\textbf{Environment} & \textbf{Episode 100} & \textbf{Episode 500} & \textbf{Episode 1000} & \textbf{Episode 2000} & \textbf{Episode 3500} \\ \hline
Breakout & 2.56 & 3.63 & 4.62 & 4.51 & 4.89 \\ \hline
\end{tabular}
\end{table}

\subsection{Transfer Learning}
We perform the following experiments using transfer learning by using pretrained encoder extracted from a pretrained model. We do this by on similar action Atari games by usually picking the ones with discrete action space of 18.

\subsubsection{Within-Game transfer learning with pretrained encoder} 
The experimental results for two different games, Breakout-v0 and Assault-v5, demonstrate that training an agent on a pretrained encoder of the same game can improve the performance of the agent in terms of rewards and duration. The experiment with DQN training from scratch took a lot of time and thus this way of transfer learning can further enhance its performance in a short time.

\begin{figure}[ht!]
  \centering
  \subfigure[Rewards vs Episode]{\includegraphics[width=0.45\textwidth]{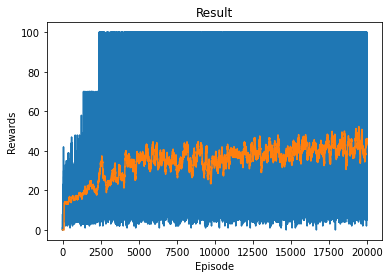}}
  \hfill
  \subfigure[Duration vs Episode]{\includegraphics[width=0.45\textwidth]{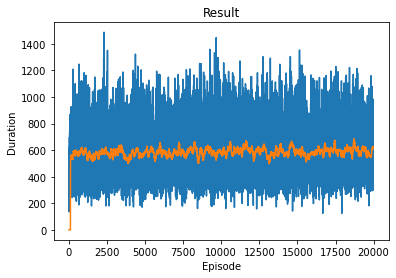}}
  \caption{Breakout on Breakout pre-trained encoder}
  \label{fig:bpp_bpE}
\end{figure}

\begin{table}[ht!]
\centering
\caption{Breakout on Breakout pre-trained encoder: Rewards obtained at specific episodes for each experimental condition.}
\label{tab:episode_rewards}
\begin{tabular}{|c|c|c|c|c|c|c|c|c|}
\hline
\textbf{Environment} & \textbf{Ep 100} & \textbf{Ep 500} & \textbf{Ep 1000} & \textbf{Ep 2000} & \textbf{Ep 5000} & \textbf{Ep 10000} & \textbf{Ep 15000} & \textbf{Ep 20000} \\ \hline
Breakout & 12.11 & 14.98 & 15.79 & 22.26 & 34.92 & 32.81 & 39.44 & 46.16 \\ \hline
\end{tabular}
\end{table}

\newpage


\begin{figure}[ht!]
  \centering
  \begin{minipage}{0.88\textwidth}
  \centering
    \includegraphics[width=\linewidth]{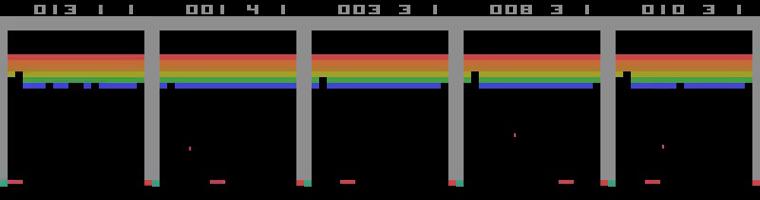}
    \label{fig:row1}
  \end{minipage}
  \begin{minipage}{0.88\textwidth}
    \centering
    \includegraphics[width=\linewidth]{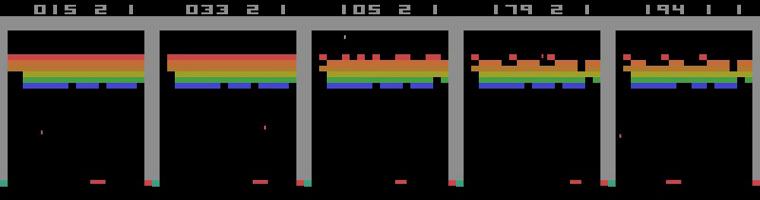}
    \caption{Breakout played using within-game transfer learning}
    \label{fig:breakout_TL}
  \end{minipage}
  \label{fig:main}
\end{figure}

\begin{figure}[ht!]
  \centering
  \subfigure[Rewards vs Episode]{\includegraphics[width=0.45\textwidth]{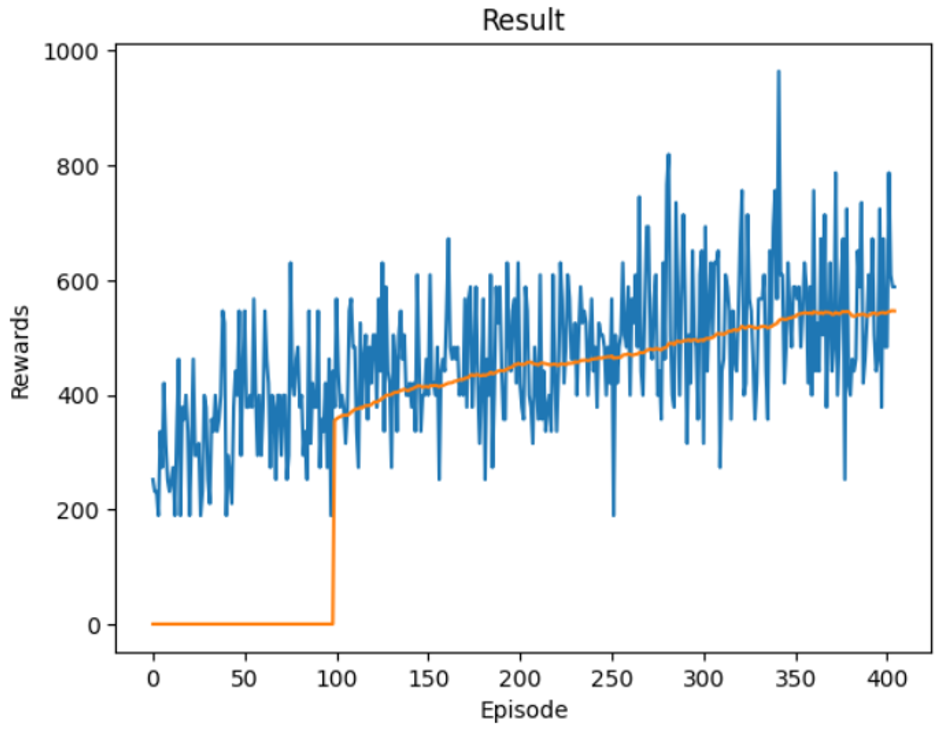}}
  \hfill
  \subfigure[Duration vs Episode]{\includegraphics[width=0.45\textwidth]{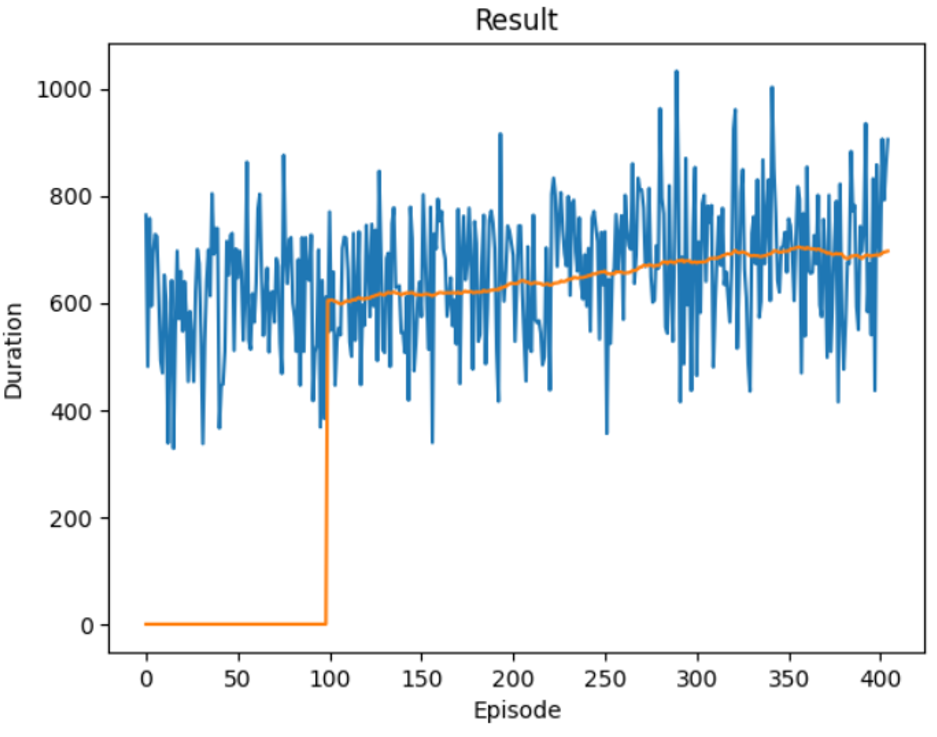}}
  \subfigure[Loss vs Episode]{\includegraphics[width=0.45\textwidth]{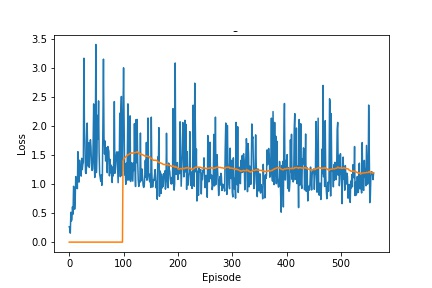}}
  \caption{Assault on assault pre-trained encoder}
  \label{fig:asp_ase}
\end{figure}

\begin{figure}[ht!]
  \centering
  \subfigure[Rewards vs Episode]{\includegraphics[width=0.45\textwidth]{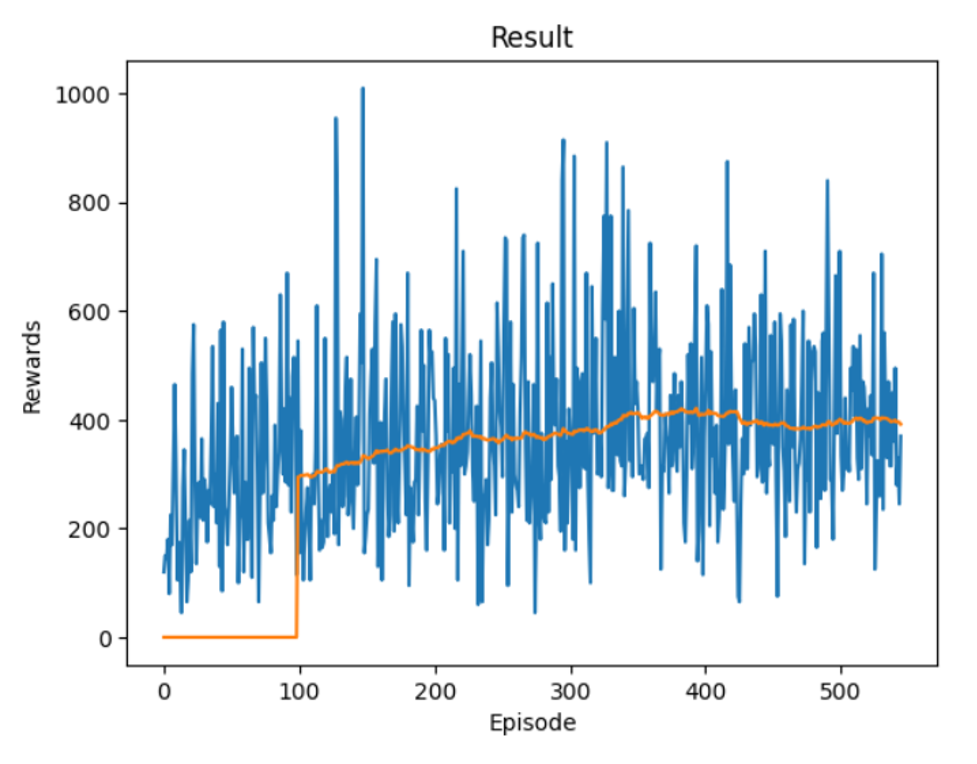}}
  \hfill
  \subfigure[Duration vs Episode]{\includegraphics[width=0.45\textwidth]{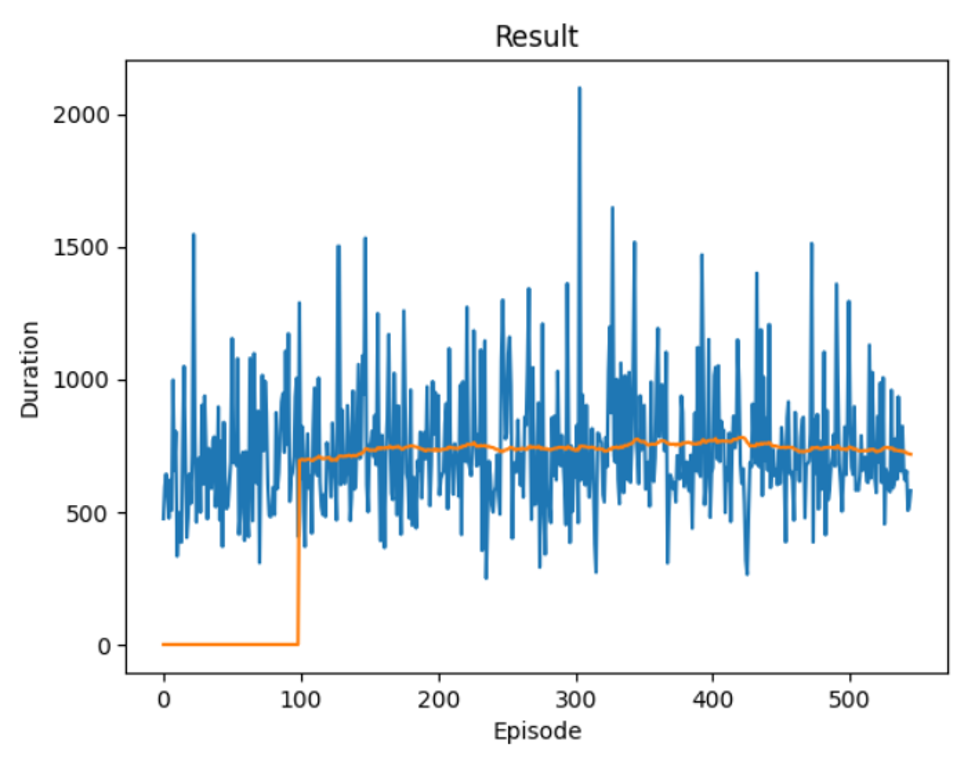}}
  \subfigure[Loss vs Episode]{\includegraphics[width=0.45\textwidth]{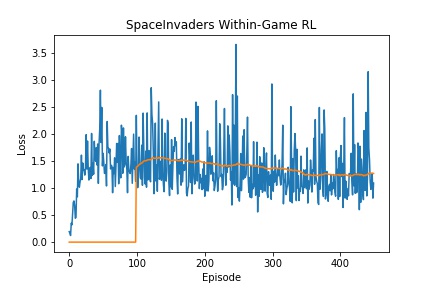}}
  \caption{Space Invaders on Space Invaders pre-trained encoder}
  \label{fig:sip_sie}
\end{figure}

\newpage
Figure \ref{fig:bpp_bpE}, \ref{fig:asp_ase} and \ref{fig:sip_sie} show the results with Rewards vs Episodes and Duration vs Episodes for each of the games. Figure \ref{fig:bpp_bpE}  shows the results of Breakout-v0 Atari environment when trained on a pretrained encoder of Breakout-v0. As we can see, the average rewards after 5000 episodes lie in the range of 36 and the average duration taken to run each episode is about 600 seconds. Additionally, from Figure \ref{fig:breakout_TL} we observe that the agent is able to learn to break through the blocks and strike high-reward blocks.

\begin{table}[ht!]
\centering
\caption{Assault on Assault pre-trained encoder: Rewards obtained at specific episodes for each experimental condition.}
\label{tab:as_on_as_episode_rewards}
\begin{tabular}{|c|c|c|c|c|c|}
\hline
\textbf{Environment} & \textbf{Ep 100} &  \textbf{Ep 150} & \textbf{Ep 200} & \textbf{Ep 300} & \textbf{Ep 400}  \\ \hline
Assault & 381.34 & 403.19 & 433.45 & 497.91 & 533.42 \\ \hline
\end{tabular}
\end{table}

Similarly, figure \ref{fig:asp_ase} shows the results of Assault-v5 environment when trained on a pretrained encoder of Assault-v5. The graphs show that the rewards start increasing after 100 episodes starting from an average of 400 up till an average value of 580. The average duration for each episode also stayed around 600 to 660 seconds on an average. The encoder used here was not from same version of the game but a previous one. Fine tuning the agent resulted in increased performance in a short span of time. We also see from figure \ref{fig:asp_ase}(c) Loss vs Episode graph, that the loss is continuously decreasing from 1.5 to 1.3 as we train more and more.

\begin{table}[ht!]
\centering
\caption{Space Invaders on Space Invaders pre-trained encoder: Rewards obtained at specific episodes for each experimental condition.}
\label{tab:si_on_si_episode_rewards}
\begin{tabular}{|c|c|c|c|c|c|}
\hline
\textbf{Environment} & \textbf{Ep 100} & \textbf{Ep 200} & \textbf{Ep 300} & \textbf{Ep 400} & \textbf{Ep 500} \\ \hline
Space Invaders & 297.12 & 377.42 & 383.19 & 402.36 & 401.02 \\ \hline
\end{tabular}
\end{table}

\begin{figure}[ht!]
  \centering
  \begin{minipage}{\textwidth}
    \centering
    \includegraphics[width=0.88\textwidth]{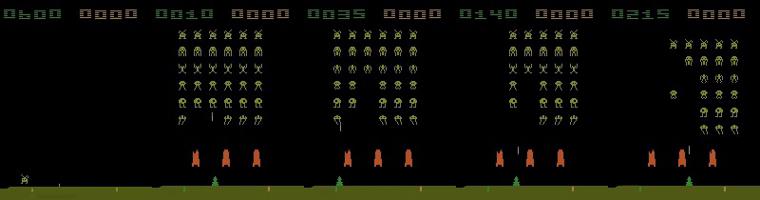}
    \label{fig:row1si}
  \end{minipage}
  \begin{minipage}{\textwidth}
    \centering
    \includegraphics[width=0.88\textwidth]{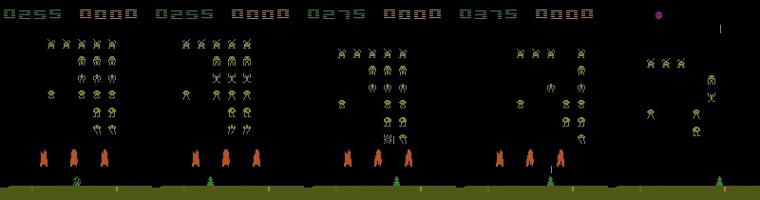}
    \caption{Space Invaders using within-game transfer learning}
    \label{fig:space_invader_within_game_TL}
  \end{minipage}
\end{figure}

Additionally, figure \ref{fig:sip_sie} displays the outcomes of using a pretrained encoder of Space Invaders for the training of an agent in Space Invaders  environment. The graphs demonstrate that the rewards started to increase after 100 episodes with an average value of 390, reaching up to an average value of 420. The average duration per episode remained steady at around 600 seconds. The encoder utilized in this experiment was from a previous version of the same game. We also see that the loss continuously decreases in the Loss vs Episode graph from 1.6 to 1.3. Also as we see in Figure \ref{fig:space_invader_within_game_TL}, the agent performs quite well against its opponents and is able to eliminate most of them with good rewards.


\subsubsection{Cross-game transfer learning with pretrained encoder} 

In reinforcement learning, transfer learning is an important technique that enables agents to leverage knowledge gained from previous tasks or environments to accelerate learning in new tasks or environments. One approach to transfer learning in RL is to use a pretrained encoder from one environment to extract features from the observations in another environment.

\begin{figure}[ht!]
  \centering
  \subfigure[Rewards vs Episode]{\includegraphics[width=0.45\textwidth]{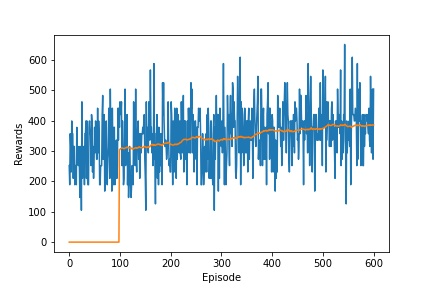}}
  \hfill
  \subfigure[Duration vs Episode]{\includegraphics[width=0.45\textwidth]{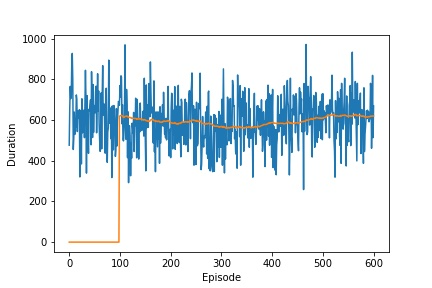}}
  \subfigure[Loss vs Episode]{\includegraphics[width=0.45\textwidth]{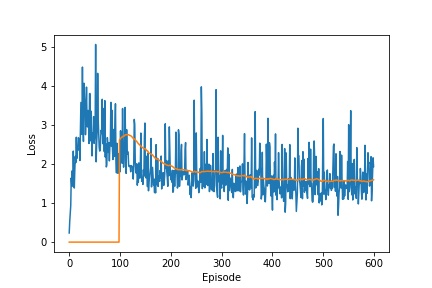}}
  \caption{Assault on Space Invaders pretrained encoder}
  \label{fig:sip_ase}
\end{figure}

Figures  \ref{fig:sip_ase} show the results of an experiment where a pretrained encoder from a different Atari game is used to extract features from the observations in another similar Atari game. This is an interesting experiment because it demonstrates that transfer learning can be effective even when the two environments are not identical. The use of a pretrained encoder in this case can be seen as a way to transfer knowledge about the structure of the game from one environment to another, which can be used to improve the learning process in the target environment.

These results demonstrate the potential of using transfer learning for cross-game scenarios, where a pretrained encoder can be leveraged to improve the performance of an agent in a different game. This approach can save time and computational resources that would otherwise be required to train a new encoder from scratch for each game.

\begin{table}[h]
\centering
\caption{Assault on Space Invaders pre-trained encoder: Rewards obtained at specific episodes for each experimental condition.}
\label{tab:sip_ase_episode_rewards}
\begin{tabular}{|c|c|c|c|c|c|c|}
\hline
\textbf{Environment} & \textbf{Ep 100} & \textbf{Ep 200} & \textbf{Ep 300} & \textbf{Ep 400} & \textbf{Ep 500}  & \textbf{Ep 600} \\ \hline
Assault & 301.62 & 312.55 & 319.67  & 361.20 & 367.52 & 384.10 \\ \hline
\end{tabular}
\end{table}

Figure \ref{fig:sip_ase} depicts an experiment on cross-game transfer learning where a pretrained encoder of the Space Invaders environment is used to train an agent on the Assault-v5 Atari game. The results show that the average reward range is about 300 to 350, with an average duration of 600 seconds per episode. This experiment demonstrates the effectiveness of transfer learning in RL for different Atari games. 

\subsubsection{Within and cross-game transfer learning, changing the softmax layer} 

\begin{figure}[t]
  \centering
  \subfigure[Rewards vs Episode]{\includegraphics[width=0.30\textwidth]{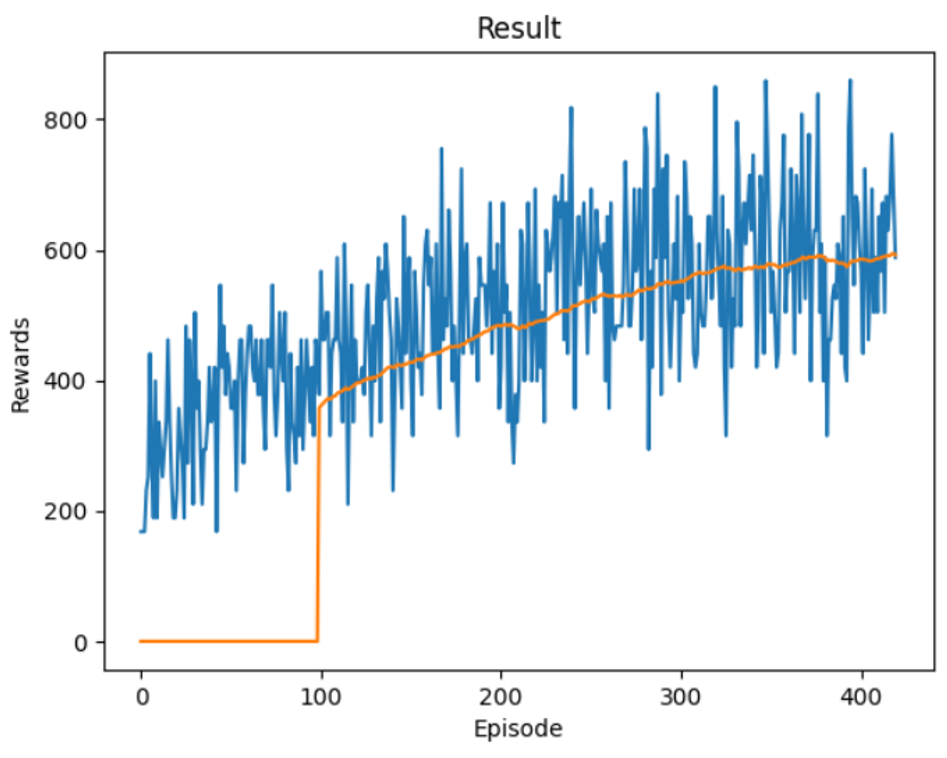}}
  \hspace{0.05\textwidth}
  \subfigure[Duration vs Episode]{\includegraphics[width=0.30\textwidth]{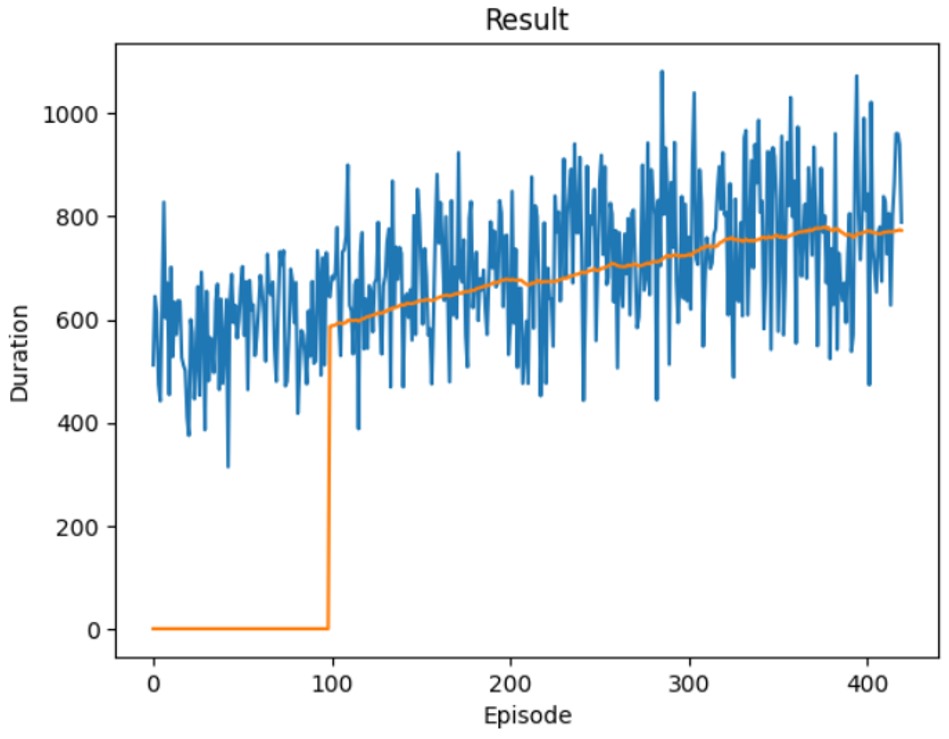}}
  \hspace{0.05\textwidth}
  \subfigure[Loss vs Episode]{\includegraphics[width=0.30\textwidth]{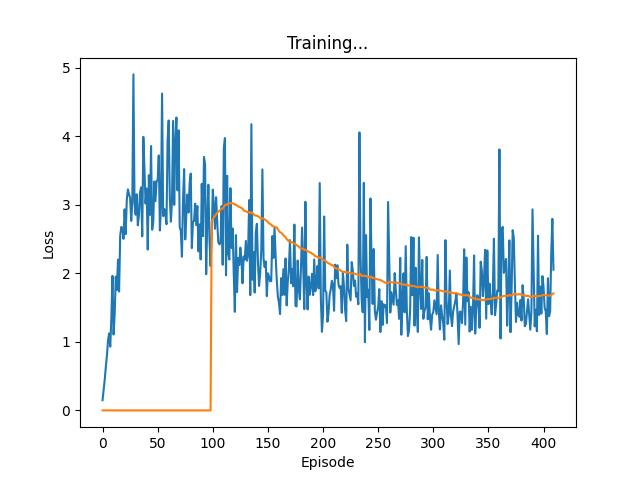}}
  \caption{Assault on assault pretrained encoder but only last linear layer modified.}
  \label{fig:asp_ase_last_linear_dur}
\end{figure}

\begin{figure}[b]
  \centering
  \subfigure[Rewards vs Episode]{\includegraphics[width=0.45\textwidth]{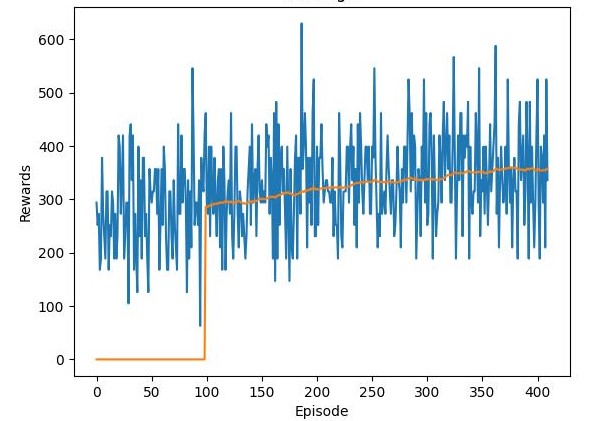}}
  \hfill
  \subfigure[Duration vs Episode]{\includegraphics[width=0.45\textwidth]{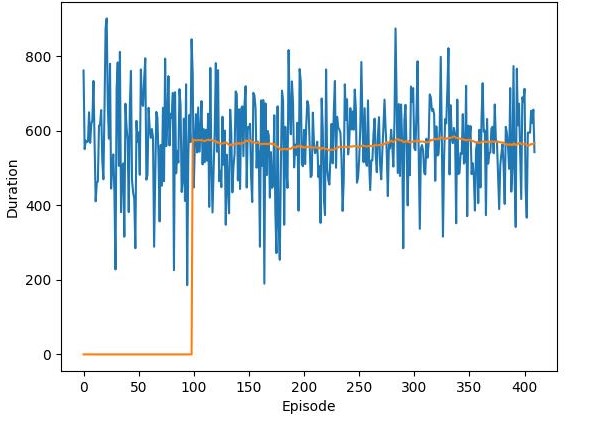}}
  \subfigure[Loss vs Episode]{\includegraphics[width=0.45\textwidth]{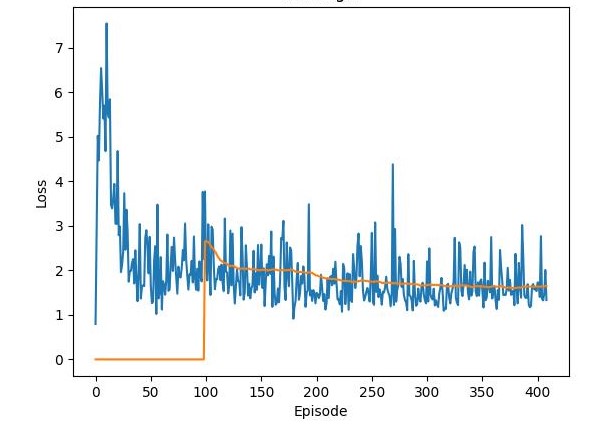}}
  \caption{Assault on Space Invader pretrained encoder but with only last linear layer modified and head initialized on
Pretrained encoder weights }
  \label{fig:sip_ase_llc_hipmw} 
\end{figure}

Figure \ref{fig:asp_ase_last_linear_dur} and \ref{fig:sip_ase_llc_hipmw} show the affect of transfer learning with only last neural network layer modified. The result in figure \ref{fig:asp_ase_last_linear_dur} shows a tremendous increase in rewards from an average of 400 to an average of 600 in just 400 more episodes. However, the training duration per episode also increased on an average.

\begin{table}[h!]
\centering
\caption{Assault on assault pretrained encoder but only last linear layer modified: Rewards obtained at specific episodes for each experimental condition.}
\label{tab:asp_ase_last_linear_dur_episode_rewards}
\begin{tabular}{|c|c|c|c|c|}
\hline
\textbf{Environment} & \textbf{Ep 100} & \textbf{Ep 200} & \textbf{Ep 300} & \textbf{Ep 400} \\ \hline
Assault & 396.43 & 472.21 & 573.89 & 588.94 \\ \hline
\end{tabular}
\end{table}

\begin{table}[ht!]
\centering
\caption{Assault on Space Invader pretrained encoder but with only last linear layer modified: Rewards obtained at specific episodes for each experimental condition.}
\label{tab:sip_ase_llc_hipmw_episode_rewards}
\begin{tabular}{|c|c|c|c|c|c|}
\hline
\textbf{Environment} & \textbf{Ep 100} & \textbf{Ep 200} & \textbf{Ep 300} & \textbf{Ep 400} & \textbf{Ep 500} \\ \hline
Space Invaders & 317.31 & 346.28 & 350.91 & 329.66 & 368.01\\ \hline
\end{tabular}
\end{table}

Figure \ref{fig:sip_ase_llc_hipmw} demonstrates the performance with an average of 350 rewards and an almost average constant duration of 600 seconds per episode.


\clearpage
\subsection{End-to-End training}
This section describes the results for plots related to the end to end training for the within as well as cross game setting. The encoder as well as the policy network is taken from the pre-trained network (encoder as well as the policy network) and initialized with its weights. The whole network is then trained of the atari game environment, Assault for our experiment. 

\begin{figure}[h!]
  \centering
  \subfigure[Rewards vs Episode]{\includegraphics[width=0.28\textwidth]{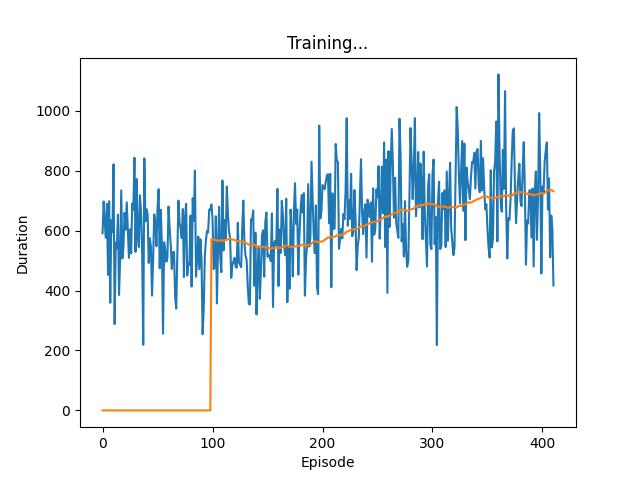}}
  \subfigure[Duration vs Episode]{\includegraphics[width=0.28\textwidth]{results/Transfer_learning/end_to_end/assault_on_assaunt/durations.jpg}}
  \subfigure[Loss vs Episode]{\includegraphics[width=0.28\textwidth]{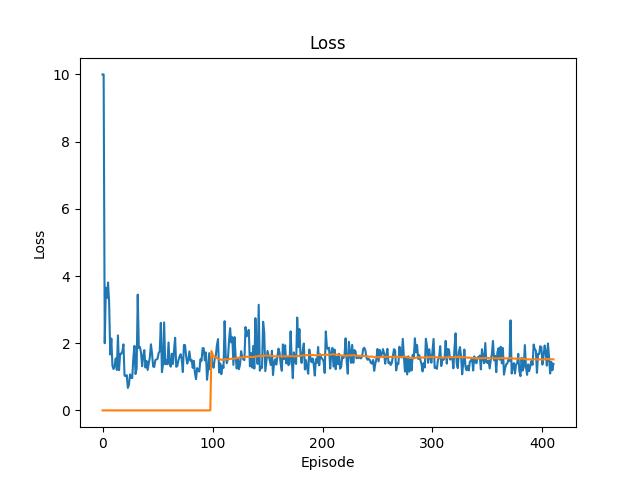}}
  \caption{Assault on Space Invaders end to end training}
  \label{fig:ete_aoa}
\end{figure}

The \ref{fig:ete_aoa}, provides plot of Reward, duration as well as loss with corresponding episodes for end-to-end Assault trained on Assault pre-trained weights, while also showing the moving average with the yellow line. Since the moving average is taken across 100 episodes so we do not have the moving average for the first 100 episodes.

\begin{figure}[ht!]
  \centering
  \subfigure[Rewards vs Episode]{\includegraphics[width=0.45\textwidth]{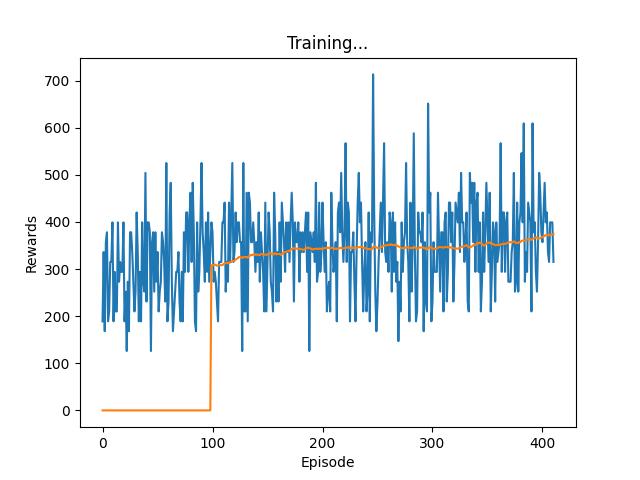}}
  \hfill
  \subfigure[Duration vs Episode]{\includegraphics[width=0.45\textwidth]{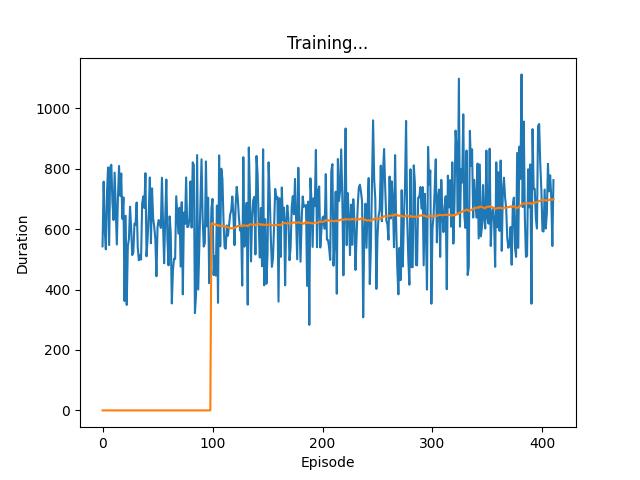}}
  \subfigure[Loss vs Episode]{\includegraphics[width=0.45\textwidth]{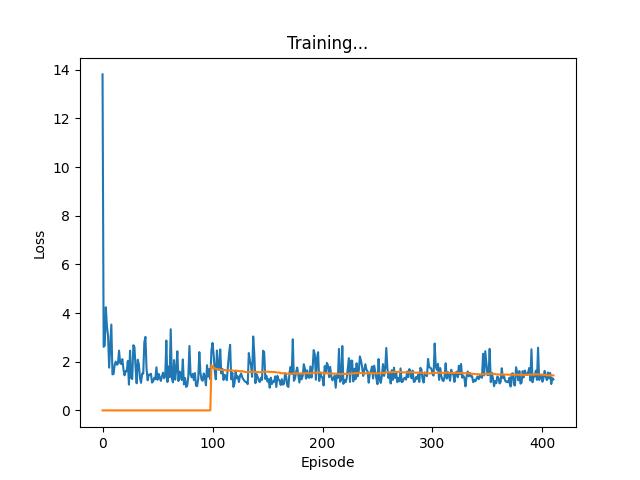}}
  \caption{Assault on Space Invaders end to end training}
  \label{fig:ete_ase}
\end{figure}

The \ref{fig:ete_ase}, provides plot of Reward, duration as well as loss with corresponding episodes for end-to-end Assault trained on Space Invader pre-trained weights, while also showing the moving average with the yellow line. Since the moving average is taken across 100 episodes so we do not have the moving average for the first 100 episodes.

\subsection{Universal Agent}
\begin{figure}[h!]
  \centering
  \begin{minipage}{\textwidth}
    \includegraphics[width=\textwidth]{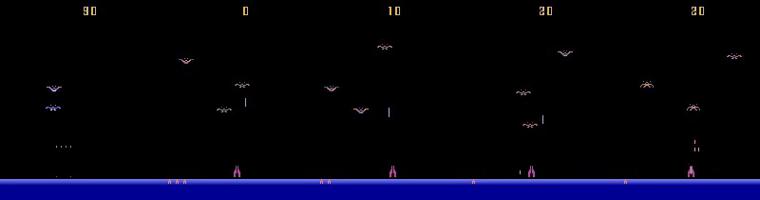}
    \label{fig:row1}
  \end{minipage}
  \begin{minipage}{\textwidth}
    \includegraphics[width=\textwidth]{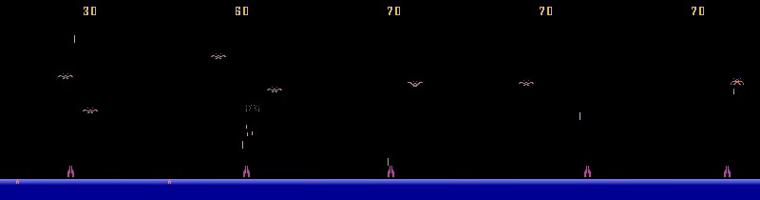}
    \caption{Demon Attack results using Universal agent}
    \label{fig:universal_agent_demon_attack}
  \end{minipage}
\end{figure}

As we see from Figure \ref{fig:universal_agent_demon_attack}, the result shows the performance on Demon Attack Atari games when the universal agent is capable of playing multiple games without prior knowledge. We can see that the game is played decently well by the agent which shoots all its attackers towards the end frames. Since this is a new unseen environment and the agent was never trained on it, we do not have the loss or the reward plots. But looking at the video, and the figure, we can see that the agent knows that it has to escape from the enemies attacks well as attack on the enemy so as to increase rewards. It knows these actions from being trained on previous similar environment, in which the agent's action space was similar and it had to attack to get rewards, and escape the enemy's attack in order to survive. Training on multiple similar environments generalizes the policy network and since a similar action-reward mapping is needed for this (Demon Attack) game, we get a decently good performance. 

When compared to the performance of agents trained on within and cross-game pre-trained encoder, we find that the universal agent's has a lower performance in terms of rewards it achieves as well as duration and achieves a mean reward of 300 for 100 game plays, while the Demon Attack train using our best method (Demon Attack on Demon Attack, using pre-trained encoder, replacing the softmax layer) achieves a mean reward of 800.

\section{Discussion}

\subsection{DQN Model from scratch}
From Figures \ref{fig:breakout_scratch_rewards}, \ref{fig:dqn_assault_scratch}, we observe that the DQN models we train on breakout and assault games do not learn significantly, even if they're trained on 15k+ episodes (where each episode constitutes multiple iterations as per the number of frames in the episode). DQN does not learn anything for breakout and the reward stays $\leq$ 5, which is a bad performance for this game. The mean reward even decreases during training, This is because when training from scratch, the CNN parameters need to be learned as well. Learning these CNN parameters is hard through these rewards and hence it takes many episodes and days for the model to start learning. This is what we attempt to tackle in our paper in the further sections through the use of pre-trained encoders.


\subsection{Reinforcement Learning - Transfer learning}
In this section we explore different approaches to apply transfer learning so as to reduce the training time as well as the respective experiment's model performance, graphs and results. The performance is measured in terms of rewards and duration. Duration is the time in one play, that the agent manages to keep playing before the game ends, whereas reward is the total reward that the agent accumulates over the duration, while playing the game. The plot for loss, shows the loss corresponding to each episode, while training.

\subsubsection{Within-Game transfer learning with pretrained encoder}

Figure \ref{fig:bpp_bpE}, \ref{fig:asp_ase} and \ref{fig:sip_sie} show the results with Rewards vs Episodes, Duration vs Episodes and loss vs Episode for each of the games, when the agent was fine-tuned using CNN encoder of the same environment but from a previous version of the environment. For the breakout agent we can see that after over 17000 episodes, the breakout agent did have a significant gain in performance. The final reward was close to 40, while the duration was close to 600. This can be due to the fact that the agent was trapped in a loop, which is one of the significant problems with breakout agents. Whereas for the assault agent, we can see a significant gain in performance in terms of reward  as well as duration in a very short span of training(400 epochs, 2 hours of training). This shows that if there is a data drift or if the model performance has decreased over time, it requires fairly less compute as well as time intensive to fine-tune a model. We also observe a decrease in the Loss vs Episodes curve where the loss keeps on decreasing continuously indicating a better performance of the agents. Additionally, Figure \ref{fig:breakout_TL}, which illustrates the performance of breakout using 10 frames, indicates that the model is able to fairly exploit the opening created on the left to hit high reward blocks, thereby demonstrating that it has learned a model of the game.

\subsubsection{Cross-Game transfer learning with pretrained encoder}

 Figures \ref{fig:sip_ase}, display the graphs of Rewards vs Episodes, Duration vs Episodes and Loss vs Episode for the experiment with cross-game transfer learning that was explored by using a pre-trained encoder of one game on to the other. The feature extractor was frozen and the head was fine-tuned as per the game. The end reward for the Assault model is around 200, while the duration is around 600. When compared \ref{fig:asp_ase}, we see that the max reward it reached was around 400 and duration was around 600, so this shows there is room for improvement, but the model could not converge. For the Assault model, its end reward is around 400 while duration is around 600. When compared to results from Figure \ref{fig:dqn_assault_scratch}, we see that the result looks promising, and converges a lot quicker and has an upward trend. 

If we compare the SpaceInvader fine-tuned on Assault CNN encoder with that of Assault trained on SpaceInvader encoder we observed that in the first case there was no gain in performance even after 400 episodes, while the model in the second case continue to gain in performance in terms of rewards. One of the reasons might be that it needed more episodes to show improvement in performance. One important point to mention here is that the number of output nodes in the final layer(Softmax) of SpaceInvader is 6 whereas for the Assault is 7. So for Assault when fine tuned could easily adjust for the six output cases, as it had encoder that produce more fine-grained outputs, so it could easily accommodate the extra information, but on the other hand SpaceInvader when trained on Assault encoder had to retrain itself to accommodate for the extra node and hence would require more training episodes. This gives as a hypothesis that, its better to take encoders from models that are complex in nature and fine-tune it for simpler use-cases.     

\subsubsection{Within and cross-game transfer learning, changing the softmax layer}

Figures \ref{fig:asp_ase_last_linear_dur} and \ref{fig:sip_ase_llc_hipmw}, display the graphs of Rewards vs Episodes, Duration vs Episodes and Loss vs Episode for the last layer initialization(changing the softmax layer). In this, the feature extractor was frozen and the head layer of the policy network was initialized by the weights from the pre-trained encoder. For the Assault on Assault extractor we see the most gain in performance across all the experiments, with end reward going as high as 590 and duration around 800. The graph curve continues to show improvement indicating that the model could even get better. The loss curve decreases continuously. One of the reason for the performance is that the head layer is already initialized with pre-trained weights and can converge quickly. For the Atari model trained on Space Invader pretrained encoder, we see a gain in performance in terms of reward, though not as high as the previous graph, Figure \ref{fig:asp_ase_last_linear_dur}. Also, the duration, remains constant and slightly decreases at the end. The loss decreases but the rate of decrease is not as steep as that of the previous experiment. 

Within game transfer learning, with replacing softmax layer is the best performing method among all the experiments, in terms of performance as well as training time. This can greatly reduce in the training time for applications in which we already have a reinforcement learning agent, and we need to retrain the agent for a newer version of the environment.

\subsection{End-to-End training}
Figures \ref{fig:ete_aoa} and \ref{fig:ete_ase}, display the graphs of Rewards vs Episodes, Duration vs Episodes and Loss vs Episode for the end-to-end training of the network, with pre-trained weights initialization for both encoder as well as the policy network. For the Figures \ref{fig:ete_aoa}, we observe that there is a constant increase in the duration as well as the reward of the model, while the loss decreases but the rate of decrease is low. The results are promising and if there is no time-constrained then is the best way to train the network if we have a same environment available, since it fine-tunes the whole network as well as decreases the convergence time. 

For the Figures \ref{fig:ete_ase}, we observe that there is a constant increase in the duration as well as the reward of the model, while the loss decreases but the rate of increase as well as decrease is low. The results are promising and if there is no time-constraint then is the best way to train the network even in cases when the there is not a pre-trained agent available for the environment. An agent with similar type of environment, can be used to initialize the weights in the new network, which drastically reduce the training time and gives better convergence. 

\subsection{Universal Agent}
The results indicate that the universal agent was able to play the Demon Attack Atari game without prior knowledge and achieved decent performance. We can observe from the video and figure that the agent was able to escape from the enemy's attacks and attack the enemies to increase rewards, which it learned from being trained on previous similar environments. Training on multiple similar environments helped generalize the policy network, resulting in a similar action-reward mapping being utilized for the Demon Attack game. The performance of the universal agent was then compared to agents trained on within and cross-game pre-trained encoders. The universal agent achieved a lower mean reward and duration in comparison, achieving a mean reward of 900 for 100 game plays.

Future work could involve training the Demon Attack game using cross-train encoders to compare performance values. However, from the video, it is clear that there is room for improvement. It is possible that further fine-tuning of the agent's hyperparameters or network architecture could result in improved performance. Additionally, incorporating other reinforcement learning techniques, such as curiosity-driven exploration or hierarchical reinforcement learning, could further enhance the agent's abilities in navigating complex and diverse environments.



\section{Conclusion}

This paper showed how transfer learning can be more effective than training from scratch which can take upto several days of computation. The results demonstrate that transfer learning significantly reduces the training time and improves the performance of the reinforcement learning models. We also show how the pretrained weights of one can can be used for training another game in the reinforcement learning setting. The use of a pretrained encoder from a relatively  complex task leads to faster convergence and better performance on a newer task with fewer training episodes. Overall, the experiments demonstrate that these methods can significantly reduce the training time and improve the performance of reinforcement learning models in Atari games. Another benefit of using a pretrained encoder is that it can aid in hyperparameter tuning. By using a pretrained encoder, it is possible to leverage the knowledge learned from the previous task to identify the best hyperparameters for the new task more efficiently. Moreover, by using a single model trained on similar environments, it is possible to achieve decent performance on previously unseen environments. This means that instead of training a new model for each new environment, a single model can be trained on a set of similar environments, and then transferred to new environments for adaptation. This approach can save time and resources and can be particularly useful in applications where the environment is constantly changing. One on the future scope of this work lies in our universal agent which can be useful in places where we have unseen environment with less or no data and we can  use the universal agent of similar multiple games and use that on the new environment. This research has potential applications in developing more efficient and effective RL algorithms for a wide range of real-world applications. Overall this paper highlights the importance of transfer learning and exploring different RL approaches to address challenges in RL, such as convergence time.

\bibliographystyle{plain} 

\bibliography{bib}

\end{document}